\def\year{2020}\relax
\documentclass[letterpaper]{article} % DO NOT CHANGE THIS
\usepackage{aaai20}  % DO NOT CHANGE THIS
\usepackage{times}  % DO NOT CHANGE THIS
\usepackage{helvet} % DO NOT CHANGE THIS
\usepackage{courier}  % DO NOT CHANGE THIS
\usepackage[hyphens]{url}  % DO NOT CHANGE THIS
\usepackage{graphicx}
\usepackage{latexsym}
\usepackage{microtype}
\usepackage{subfig}
\usepackage{booktabs} % for professional tables
\usepackage{amsmath}

\makeatletter
\newcommand{\printfnsymbol}[1]{%
  \textsuperscript{\@fnsymbol{#1}}%
}
\makeatother

\title{TextNAS: A Neural Architecture Search Space tailored for Text Representation}
\author{
\Large \textbf{Yujing Wang\textsuperscript{\rm 1,2,\thanks{Equal Contribution}}, Yaming Yang\textsuperscript{\rm 1,\printfnsymbol{1}}, Yiren Chen\textsuperscript{\rm 2,\printfnsymbol{1},\thanks{The work was done when the authors visited MSRA}}, Jing Bai\textsuperscript{\rm 1}, Ce Zhang\textsuperscript{\rm 3,\printfnsymbol{2}}} \\
\Large \textbf{Guinan Su\textsuperscript{\rm 4,\printfnsymbol{2}}, Xiaoyu Kou\textsuperscript{\rm 2,\printfnsymbol{2}}, Yunhai Tong\textsuperscript{\rm 2}, Mao Yang\textsuperscript{\rm 1}, Lidong Zhou\textsuperscript{\rm 1}}\\ % All authors must be in the same font size and format. Use \Large and \textbf to achieve this result when breaking a line
\textsuperscript{\rm 1}Microsoft Research Asia \\
\textsuperscript{\rm 2}Key Laboratory of Machine Perception, MOE, School of EECS, Peking University \\
\textsuperscript{\rm 3}ETH Z\"{u}rich $~~~~~~$
\textsuperscript{\rm 4}University of Science and Technology of China \\
\{yujwang, yayaming, jbai, maoyang, lidongz\}@microsoft.com \\
\{yrchen92, kouxiaoyu, yhtong\}@pku.edu.cn \\
ce.zhang@inf.ethz.ch, sa517299@mail.ustc.edu.cn
}

\begin{document}

\maketitle

\begin{abstract}
Learning text representation is crucial for text classification and other language related tasks. There are a diverse set of text representation networks in the literature, and how to find the optimal one is a non-trivial problem. Recently, the emerging Neural Architecture Search (NAS) techniques have demonstrated good potential to solve the problem. Nevertheless, most of the existing works of NAS focus on the search algorithms and pay little attention to the search space. In this paper, we argue that the search space is also an important human prior to the success of NAS in different applications. Thus, we propose a novel search space tailored for text representation. Through automatic search, the discovered network architecture outperforms state-of-the-art models on various public datasets on text classification and natural language inference tasks. Furthermore, some of the design principles found in the automatic network agree well with human intuition.
\end{abstract}

%  text classification \cite{joulin2016bag}, machine translation \cite{wu2016google}, and reading comprehension \cite{yu2018qanet}
\section{Introduction}
Neural network models have demonstrated their superiority in many natural language tasks such as text classification, machine translation and reading comprehension. One of the core problems of natural language processing is to design a network architecture that effectively captures the syntax and semantics incorporated in texts. Contrary to the computer vision domain where CNN is predominant, the state-of-the-art neural networks for text representation are much more diverse, including CNN~\cite{zhang2015character}, RNN~\cite{liu2015representation}, hybrid model of CNN+RNN~\cite{zhou2015c,tang2015document} and Transformer~\cite{vaswani2017attention}, etc. Nevertheless, how to find the optimal text representation network is still an unsettled problem in the literature.

Recently, Neural Architecture Search (NAS) techniques have opened up a new opportunity for customized architecture design. Existing works of NAS mainly focus on the study of search algorithms and put little emphasis on the search space. However, there remain several challenges for applying NAS to different applications. First, it is prohibitive to search for all kinds of possibilities thoroughly, even when advanced search algorithms (for example, gradient-based, evolution, reinforcement learning, etc.) are utilized; Second, when the search space is extra-large, the NAS algorithm may select a neural architecture that overfits to both training and validation data. Thus, we argue that the search space is an indispensable human prior which deserves more investigation in different applications.

In this paper, we propose TextNAS, a novel search space customized for text representation. The search space is designed based on the following motivations and findings:
\begin{itemize}
\item \textbf{It is beneficial to explore the customized solution of layer mixture.}
It is well-known that different layers are beneficial from different perspectives. CNN is good at learning local feature combinations (analogies to n-grams), RNN specializes in sequential modeling, and Transformer \cite{vaswani2017attention} is able to capture long-distance dependencies directly. There are some evidences demonstrating the potential of layer mixture, for instance, C-LSTM \cite{zhou2015c} utilizes CNN to extract a sequence of higher-level phrase representation and then feeds the CNN output to another RNN layer to produce the ultimate sentence embedding vectors.
\item \textbf{The macro search space is a better choice for text representation}
Most previous works of NAS prefer micro search space \cite{zoph2017learning} as they work well on image-related tasks. However, according to a preliminary experiment (showed in Table \ref{table:micro_macro}), we demonstrate that the macro search space is better than the micro one in the text classification scenario. This shows the necessity of leveraging customized search spaces for different applications.
\item \textbf{The search space should support multi-path ensembles.}
One limitation of existing macro search space is that it only embodies single-path neural networks. However, multi-path ensemble is a common design principle in manual networks, e.g., InceptionV4~\cite{szegedy2017inception}. Intuitively, different categories of layers act as distinct feature extractors, an ensemble of which provides potentially better representation for the sentence.

\begin{table}[t]
    \caption{Comparison of micro and macro search spaces on different tasks using ENAS search algorithm}
    \small
    \begin{center}
    \begin{tabular}{l|c|c|c}
    \toprule
    \textbf{Dataset} & \textbf{Task} & \textbf{Acc (micro)} & \textbf{Acc (macro)}\\
    \midrule
    CIFAR10 & Image Classification & \textbf{97.11} & 95.67 \\
    SST & Text Classification & 47.00 & \textbf{51.55} \\
    YAHOO & Text Classification & 70.63 & \textbf{73.16} \\
    AMZ & Text Classification & 58.27 & \textbf{62.64} \\
    \bottomrule
    \end{tabular}
    \end{center}
\label{table:micro_macro}
\end{table}
\end{itemize}

The TextNAS search space consists of a mixture of convolutional, recurrent, pooling and self-attention layers. It is based on a general DAG structure and supports the ensemble of multiple paths. Given the search space, the TextNAS pipeline can be conducted in three procedures.\footnote{The open source code can be found at:\\ https://github.com/microsoft/nni/tree/master/examples/nas/textnas} (1) The ENAS \cite{pham2018efficient} search algorithm is performed on the search space by utilizing the evaluation accuracy on validation data as RL reward; (2) Grid search is conducted by the optimal architecture to search for the best hyper-parameter setting on the validation set. (3) The derived architecture is trained from scratch with the best hyper-parameters on the combination of training and validation data.

We ran experiments on the Stanford Sentiment Treebank (SST) dataset \cite{socher2013recursive} to evaluate the TextNAS pipeline. The experimental results showed that the automatically generated neural architectures achieved superior performances compared to manually designed networks. We look into the automatic architecture and find that some of the design principles agree well with human experiences. Moreover, since the neural architecture search procedure is time- and resource-consuming, we are interested in the transferability of the derived network architectures to other text-related tasks. Impressively, the transferred architectures outperformed current state-of-the-art methods \cite{zhang2015character,yang2016hierarchical,conneau2016very} on various text classification and natural language inference datasets.

%To the best of our knowledge, our work is the \textit{first attempt} that leverages \textit{neural architecture search} to design a neural network specialized for text representation. Through automatic search, we discover novel layer assemblings that have not been studied by human experts before. We thus advocate more research on the enhancement of architecture search spaces tailored for NLP applications, which is orthogonal to the optimization of search algorithms. The derived architectures also show good transferability, so we can construct neural networks for large-scale datasets efficiently by searching on smaller datasets.

\section{Related Work}
\subsection{Neural Architecture Search}
Neural Architecture Search (NAS) has become an important research topic in AutoML domain, the goal of which is to find the optimal network structure in a given search space which achieves excellent performance on a specific task. Existing studies in this direction can be summarized in two aspects. One line of research focuses on evolution algorithms, which offer flexible approaches for generating neural networks by simultaneously evolving along network structures and hyper-parameters \cite{real2018regularized}. Another line of research concentrates on reinforcement learning, for example, NAS (Neural Architecture Search) \cite{zoph2016neural} leverages a recurrent neural network as controller to generate child networks, while the controller is trained with reinforcement learning. Despite of impressive performance, the original NAS framework is computationally expensive.

There are various attempts to improve the search efficiency of NAS. \cite{zoph2017learning} reduces the search space to two micro cells: the normal cell and the reduction cell, while the cells can be stacked to construct deep neural networks; PNAS \cite{liu2017progressive} adopts a sequential model-based optimization strategy and constructs the network layer by layer while simultaneously learns a surrogate model to guide the search routine; \cite{baker2017accelerating} accelerates the search procedure through predicting the final performance by partially trained model configurations; ENAS \cite{pham2018efficient} accelerates the reinforcement learning procedure by sharing parameters among child trials; DARTS \cite{liu2018darts} formulates the task of neural architecture search in a differentiable manner and does not require reinforcement learning controllers; SMASH \cite{brock2017smash} proposes one-shot model architecture search by designing a hyper-network to generate the parameter values for each model; \cite{bender2018understanding} demonstrates the possibility of leveraging one-shot architecture search to identify promising architectures without hyper-networks or reinforcement learning; \cite{li2019random} shows that random search with early-stop is a competitive NAS baseline and random search with weight-sharing achieves further improvement.

\subsection{Text Classification}

RNN is specialized for long sequential modeling and has the capability of processing variable-length inputs, making it a natural choice for text classification. For example, \cite{tai2015improved} introduces a tree-structured LSTM network to capture sentence meanings with emphasis on the syntactic structure. At the same time, there is another branch of methods using CNN for text classification \cite{dos2014deep,zhang2015character,conneau2016very}. Benefit from the advantages of both RNN and CNN, there is a growing interest in assembling them, including C-LSTM \cite{zhou2015c}, RCNN \cite{kalchbrenner2013recurrent} and GatedNN \cite{tang2015document}. These models utilize CNN to extract a sequence of higher-level phrase representation and feed the CNN output to additional RNN layers to produce the ultimate text representation vectors. Moreover, attention mechanism \cite{luong2015effective} has been widely adopted in NLP applications, which enables neural networks to focus on specific parts in the text sequence. As an example, \cite{yang2016hierarchical} proposes a hierarchical attention network where two attention layers are applied at word and sentence level respectively. In addition, Transformer \cite{vaswani2017attention} invents multi-head self-attention in the text encoder to relate different positions of a single word sequence.

\subsection{Natural Language Inference}
Natural Language Inference (NLI) is another fundamental NLP task that determines the inferential relationship among sentences. There are two major categories of neural network models for NLI, namely sentence vector-based models and joint models. The former represents each sentence as a fixed-length vector before inferring the relationship between them; while the latter utilizes cross-sentence layers explicitly in the neural network for relation prediction. In this paper, the goal is to evaluate the capability of text representation, so we adopt the sentence-vector based framework. Conneau et al. \cite{conneau2017supervised} compared 7 different network architectures and showed that a single BiLSTM layer with max pooling can act as the universal sentence encoding model. Based on this work, \cite{nie2017shortcut} designed a stacked BiLSTM layer with shortcut connections and \cite{talman2018hbmp} devised a hierarchical BiLSTM max pooling (HBMP) model. Besides, \cite{chen2018enhancing} proposed a new vector-based multi-head attention pooling layer to enhance the sentence representation; \cite{im2017distance} utilized the self-attention network that considered local dependencies of different words to generate distance-based sentence embedding vectors; \cite{yoon2018dynamic} combined the self-attention mechanism with modified dynamic routing borrowed from the capsule network.

\section{TextNAS}

\begin{figure}
\centering
\subfloat[]{
  \includegraphics[width=0.2 \textwidth]{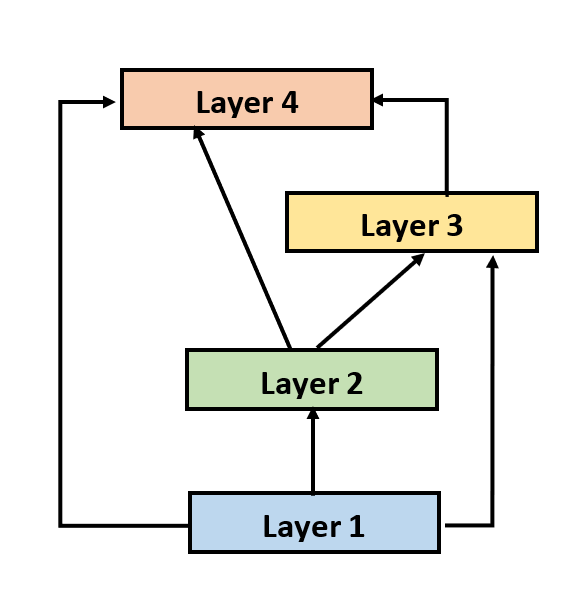}
  \label{fig:DAG_search_space}
}
\subfloat[]{
  \includegraphics[width=0.2 \textwidth]{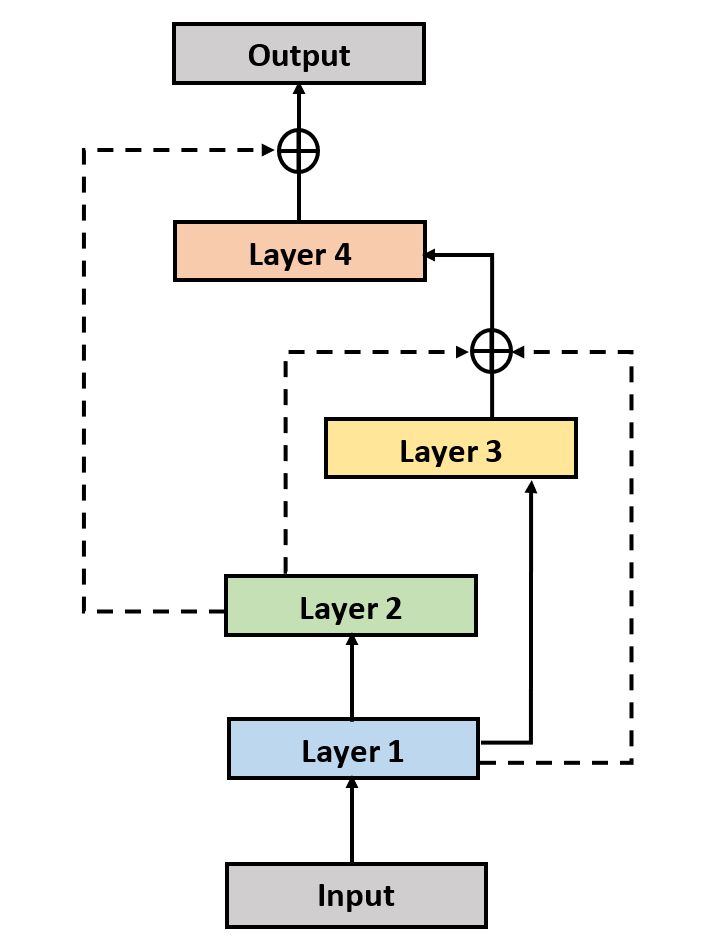}
  \label{fig:DAG_instance}
}
\caption{(a) The general DAG search space of four layers (b) A neural network instance sampled from the general search space.}
\label{fig:DAG}
\end{figure}

In this section, we introduce our method in details. First, we propose the novel search space tailored for text representation. Second, we introduce the search algorithms adopted in TextNAS. Finally, we describe the frameworks of two tasks, i.e., text classification and natural language inference.

\subsection{Search Space}
\label{search_space}
%In natural language applications, the neural networks often consist of multiple categories of layers, including convolutional, recurrent, pooling, and multi-head self-attention layers. As different layers are designed for different purpose, we sometimes get insights of how to assemble them as a deep architecture for performance enhancement. For example, CNN is known to be good at local feature combination and RNN is specialized for sequential modeling. Based on this insight, \cite{zhou2015c} proposes a neural network that leverages convolutional layers in the bottom to produce local features (analogies to n-grams), while recurrent layers are applied on the top to model long-term dependencies in the word sequence. Although human experts have several successful trials of layer assembling, it is exhaustive to try all combinations manually and many possibilities have not been explored before. In this paper, our philosophy is to design a general search space for layer assembling and resort to an automatic algorithm to search for the best text representation architecture.

The macro search space of neural network can be depicted by a general DAG. As shown in Figure \ref{fig:DAG_search_space}, every node in the DAG represents a layer, and every edge from node $i$ to node $j$ denotes that layer $i$ is served as an input or skip-connection to layer $j$. Without loss of generality, we define a topological order for the layers, where layer $0$ stands for the original input layer and an edge $<$$i$, $j$$>$ exists when $i < j$. Based on the DAG search space, a network instance can be sampled by traversing the layers according to the topological order. For each layer $i$, we first choose a unique input layer from one of the previous layers $\{0, 1, ..., i-1\}$; then we make multiple choices from previous layers as skip connections, which are summed with the output of layer $i$. An example of the network instance is shown in Figure \ref{fig:DAG_instance}, which can be generated in the following steps: (1) layer $2$ and $3$ both choose layer $1$ as input; (2) layer $3$ chooses layer $1$ and $2$ as additional skip connections (shown in dotted lines); (3) layer $4$ chooses layer $3$ as input and layer $2$ as an additional skip connection.

\begin{figure}
\centering
\subfloat[]{
  \includegraphics[width=0.2 \textwidth]{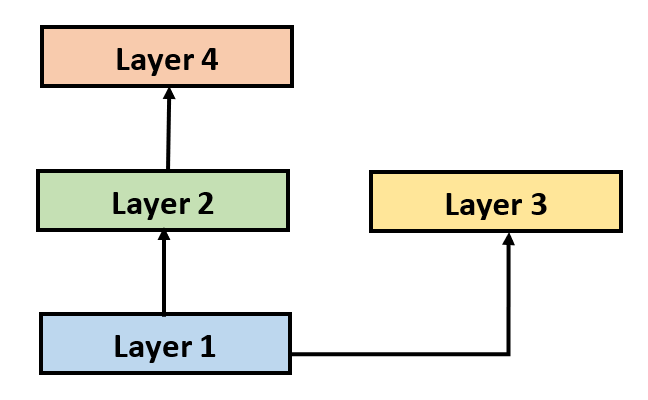}
  \label{fig:duplicate_a}
}
\subfloat[]{
  \includegraphics[width=0.2 \textwidth]{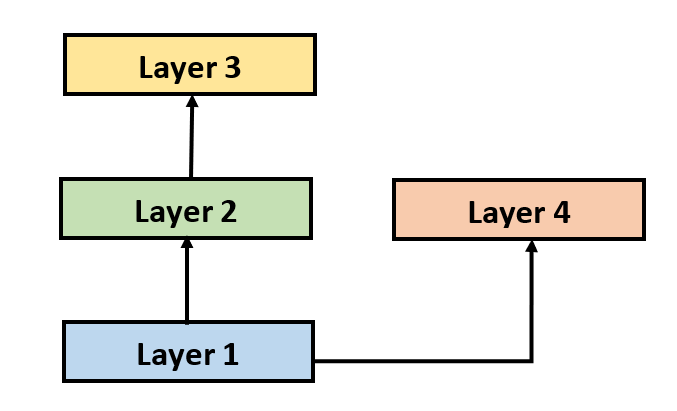}
  \label{fig:duplicate_b}
}
\caption{Duplicated network exmaples constructed by different orders}
\label{fig:duplicate}
\end{figure}

We notice that different construction orders sometimes lead to the same network architecture, as illustrated in Figure \ref{fig:duplicate}. We put a constraint on the search space to mitigate this kind of duplication and accelerate the search procedure. Concretely, layer $i$ must select its input from previous $k$ layers, where $k$ is set to be a small value. In this way, we favor the BFS-style construction manner in Figure \ref{fig:duplicate_a} instead of Figure \ref{fig:duplicate_b}. For example, if we set $k = 2$, the case in Figure \ref{fig:duplicate_b} can be skipped because layer $4$ cannot take layer $1$ as input directly. In our experiments, we set $k=5$ as a trade-off between expressiveness and search efficiency.

The tensor shape of the input word sequence is $<$$batch\_size$, $emb\_dim$, $max\_len$$>$, where $batch\_size$ is the pre-defined size of mini-batch; $emb\_dim$ is the embedding dimension of word vectors and $max\_len$ denotes the max length of the word sequence. In our implementation, we adopt a fixed-length representation, i.e., additional $pad$ symbols are added to the tail if the input length is smaller than $max\_len$; and the remaining text is discarded if the input length is larger than $max\_len$. In all the layers, we keep the tensor shape as $<$$batch\_size$, $dim$, $max\_len$$>$, where $dim$ is the dimension of hidden units. Note that $dim$ may not equal to $emb\_dim$, so an additional 1-D convolution layer is applied after the input layer.

After the network structure is built, the next step is to determine the options for each layer. In the search space, we incorporate four categories of candidate layers which are commonly used for text representation, namely \textit{Convolutional Layers}, \textit{Recurrent Layers}, \textit{Pooling Layers}, and \textit{Multi-Head Self-Attention Layers}. Each layer does not change the shape of input tensor, so one can freely stack more layers as long as the input shape is not modified.

\textbf{Convolutional Layers}. We define four kinds of 1-D convolution layers as candidate options with filter size 1, 3, 5, and 7 respectively. To keep the shape of output the same as input, we utilize the convolution of $stride = 1$ with SAME padding; and the number of output filters is equal to the input dimension. Note that the 1-D convolution with $filter\_size=1$ and $stride=1$ is analogue to a feed-forward layer. We apply Relu-Conv-BatchNorm once a convolutional layer is added.

\textbf{Recurrent Layers}. There are multiple kinds of recurrent layers, e.g., the vanilla RNN \cite{horne1995experimental}, LSTM \cite{hochreiter1997long} and GRU \cite{bahdanau2014neural}. LSTM and GRU are known to be more advantageous than the vanilla RNN for capturing long-term dependencies in a text sequence; while GRU is usually several times faster than LSTM without loss of precision \cite{chung2014empirical}. Therefore, we leverage GRU layer as our RNN implementation. Specifically, we implement a bi-directional GRU that sums the output vectors of two opposite directions. One can also make LSTM and GRU as two candidate layers and let the search algorithm to make the decision.

\textbf{Pooling Layers}. The pooling layers calculate the maximum or average value within a filter window. We use pooling operations with SAME padding and $stride = 1$ so that the dimension of tensor does not change after pooling. For simplicity, we fix the filter size as 3 and only search between maximum or average pooling options. One can also enlarge the search space by allowing multiple choices of the filter size.

\textbf{Multi-Head Self-Attention Layers}. Multi-head self-attention layer is a major component in the neural network of Transformer \cite{vaswani2017attention}. A Transformer block is constructed by one multi-head self-attention layer followed by one or more feed-forward layers. In our search space, we already have analogous to feed-forward layers, so we leverage the automatic search algorithm to decide how to combine them. The number of attention heads is set as 8 in all the experiments. We do not use positional embedding for the input of multi-head self-attention layers because it will destroy the translation invariance of succeeding pooling and CNN layers.

\subsection{Search Algorithm}
\label{enas}
We leverage the ENAS (Efficient Neural Architecture Search) search algorithm \cite{pham2018efficient} because it is one of most effective and efficient among all state-of-the-art search algorithms. ENAS searches for the best network architecture via reinforcement learning with weight sharing. In each step, the controller is responsible for sampling several child networks from the general search space. Then the child architectures are trained on the training set and evaluated on the validation set. The child networks share the same set of parameters with the global super-graph to accelerate the evaluation procedure. After the performance of each child network is evaluated, the accuracy is fed back to the controller and the parameters are updated through policy gradients based on REINFORCE \cite{williams1992simple}.

We reuse the open source code\footnote{https://github.com/melodyguan/enas} of ENAS and implement the our novel search space accordingly. Concretely, the controller is implemented by a single LSTM layer, which generates the choice of each layer sequentially according to its topological order. For layer $i$, it first samples an input layer ID among $[max(0, i-k), i-1]$ via softmax probabilities. Then it generates $i$ binary outputs by sigmoid to identify if layer $0$, $1$, ..., $i - 1$ have skip connections with layer $i$. At last, an operator is selected for each layer. There are totally 8 options from 4 categories, i.e., 1-D convolution with filter size 1, 3, 5, 7; max pooling; average pooling; Gated Recurrent Units (GRU) and multi-head self-attention. The selection probabilities of these options are calculated by softmax.

\begin{figure}
    \centering
    \includegraphics[width=0.4\textwidth]{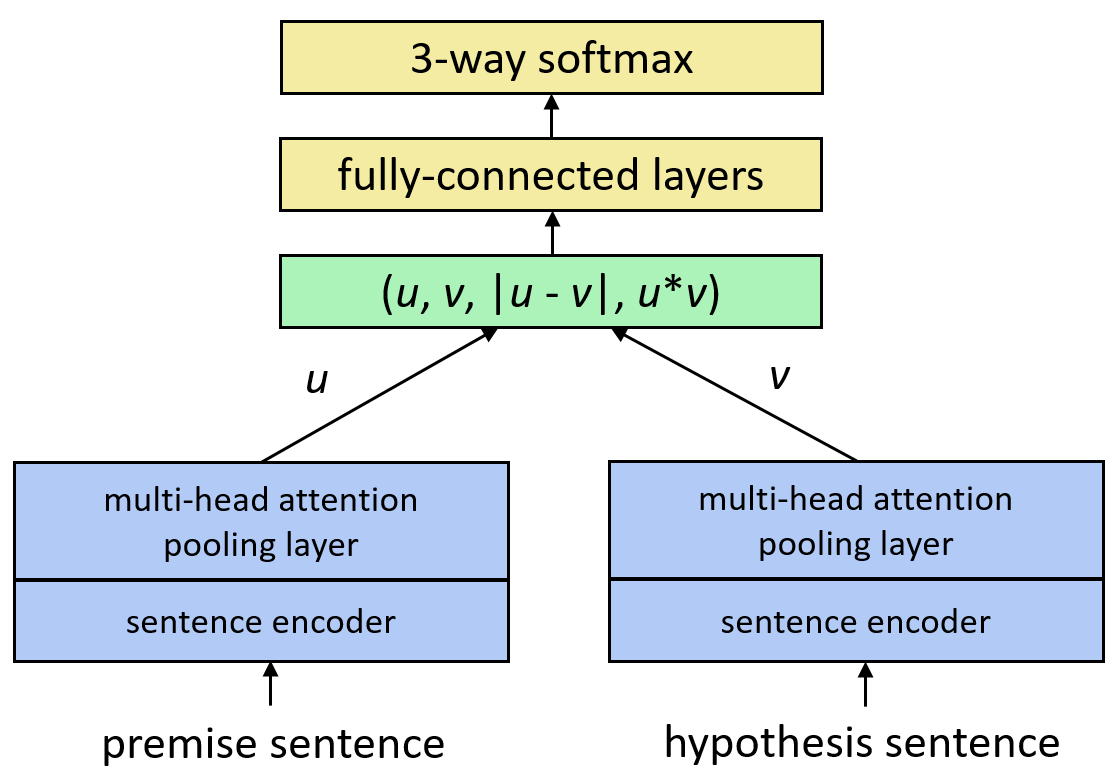}
    \caption{\label{SNLI} The sentence vector-based framework for natural language inference task}
    \vskip -0.2in
\end{figure}

\subsection{Tasks}
\label{framework}
We evaluate on two tasks to verify the feasibility and generality of our approach.

\textbf{Text Classification} is the task of assigning tags or categories to text according to its content. All layers in the text representation network are linearly combined \cite{peters2018deep} and followed by a max pooling layer and a fully connected layer with softmax activation to output the classification result.

\textbf{Natural Language Inference} is the task of determining whether a hypothesis sentence is entailment, contradiction or neutral given a premise sentence. We adopt the sentence vector-based framework \cite{bowman2015large} for this task since our goal is to compare different text representation architectures. The framework is illustrated in Figure \ref{SNLI}. The two sentences (i.e., hypothesis and premise) share the same text representation network, while the multi-head attention pooling layer \cite{chen2018enhancing} is applied on top to generate the sentence embedding vector $u$ and $v$. After that, we concatenate $u$, $v$, absolute element-wise distance $|u-v|$ and element-wise product $u \cdot v$ to construct the feature vector. We then feed the feature vector to three fully connected layers with ReLU activation before calculating 3-way softmax output.

\begin{figure*}[ht]
    \vskip 0.18in
    \begin{center}
    \centerline{\includegraphics[width=\textwidth]{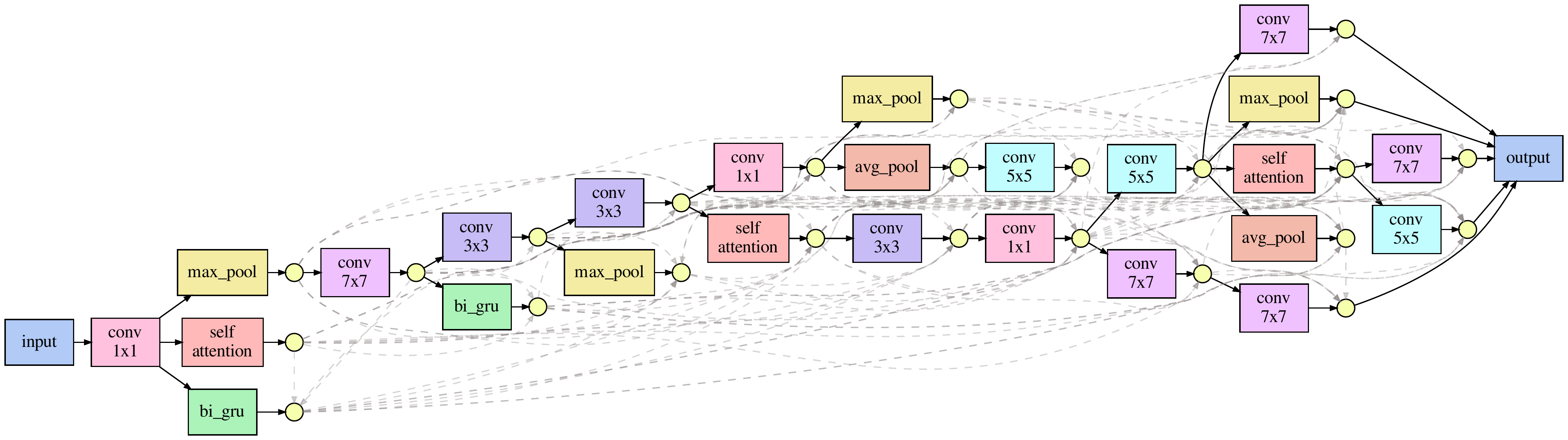}}
    \end{center}
    \caption{\label{ARC-I} Visualization of TextNAS network: Rectangles represent layers, circles represent summations, one-way arrows represent inputs, and dotted one-way arrows represent skip connections.}
\end{figure*}

\section{Experiments}
We first conduct neural architecture search and evaluate the performance on \textit{SST}, a medium size dataset of text classification which has been extensively studied by human experts. Then we transfer the derived architectures to other text classification and natural language inference tasks.

%In this section, we show the experimental results on various publicly available datasets of text classification and natural language inference tasks to verify the effectiveness and versatility of our proposed method.

%\subsection{Text Classification}
%In Section \ref{exp:nas}, we perform neural architecture search on a relatively small dataset, \textit{SST}, which has been extensively studied by human experts. Then, the top three architectures from the search results are evaluated on \textit{SST} (Section \ref{exp:sst}) and transferred to other eight text classification datasets (Section \ref{exp:transfer}). Besides, we compare the performance of different neural networks generated from single-path and multi-path DAG search spaces in Section \ref{exp:multi}.

\begin{table}[t]
    \caption{Statistics of text classification datasets}
    \vskip 0.15in
    \begin{center}
    \begin{small}
    \begin{sc}
    \begin{tabular}{lllcl}
    \toprule
    Dataset & \#Class & \#Train & \#Valid & \#Test \\
    \midrule
    SST &            5&      8,544& 1,101 & 2,210         \\
    SST-B &     2&      6,920& 872 & 1,821    \\
    AG &         4&    120,000& - & 7,600    \\
    SOGOU     &      5&    450,000& - & 60,000   \\
    DBP &       14&    560,000& - & 70,000   \\
    YELP-B &    2&    560,000& - & 38,000   \\
    YELP  &          5&    650,000& - & 50,000   \\
    YAHOO     &     10&  1,400,000& - & 60,000   \\
    AMZ      &    5&  3,000,000& - & 650,000  \\
    AMZ-B &  2&  3,600,000& - & 400,000  \\
    \bottomrule
    \end{tabular}
    \end{sc}
    \end{small}
    \end{center}
    \vskip -0.1in
\label{table:dataset}
\end{table}

\subsection{Neural Architecture Search}
\textit{SST} is short for Stanford Sentiment Treebank \cite{socher2013recursive} which is a commonly used dataset for sentiment classification. There are about 12 thousand reviews in \textit{SST} and each review is labeled to one of the five sentiment classes. There is another version of the dataset, \textit{SST-Binary}, which has only two classes representing positive/negative while the neutral samples are discarded.

%The dataset statistics are shown in Table \ref{table:dataset}.

In our experiments, we perform 24-layers neural architecture search on \textit{SST} dataset and evaluate the derived architectures on both \textit{SST} and \textit{SST-Binary} datasets. We follow the pre-defined train/validation/test split of the original datasets\footnote{https://nlp.stanford.edu/sentiment/code.html}. The word embedding vectors are initialized by pre-trained GloVe (glove.840B.300d\footnote{https://nlp.stanford.edu/projects/glove/}) \cite{pennington2014glove} and fine-tuned during training. We set the batch size as 128, max input length as 64, hidden unit dimension for each layer as 32, dropout ratio as 0.5 and $L_2$ regularization as $2 \times 10^{-6}$. We utilize Adam optimizer and learning rate decay with cosine annealing:
\begin{equation}
    \lambda = \lambda_{min} + 0.5 \cdot (\lambda_{max} - \lambda_{min})(1 + cos(\pi T_{cur} / T))
\end{equation}
where $\lambda_{max}$ and $\lambda_{min}$ define the range of the learning rate, $T_{cur}$ is the current epoch number and $T$ is the cosine cycle. In our experiments, we set $\lambda_{max}=0.005$, $\lambda_{min}=0.0001$ and $T=10$. After each epoch, ten candidate architectures are generated by the controller and evaluated on a batch of randomly selected validation samples. After training for 150 epochs, the architecture with the highest evaluation accuracy is chosen as the text representation network.

The whole process can be finished within 24 hours on a single Tesla P100 GPU. As visualized in Figure \ref{ARC-I}, the automatically discovered architecture is assembled by multiple paths and different categories of layers, including 13 convolution layers, 4 max-pooling layers, 2 average-pooling layers, 2 bi-directional GRU layers and 3 self-attention layers. Although it is much more complex than manual architectures, we still find that there are some design principles in line with human common-sense:

\begin{itemize}
\item The avg/max pooling layers and CNN/GRU/self-attention layers are alternatively stacked. The pooling layers help for extracting rotational/positional invariant features as inputs to other layers.
\item There are convolution layers before and after each GRU and multi-head self-attention layers, which is similar to C-LSTM~\cite{zhou2015c} and Transformer~\cite{vaswani2017attention}. Intuitively, convolution operations generate local feature combinations (similar to n-grams) as complementary to GRU/self-attention layers which mainly capture long-term dependencies.
\item The design principles look similar to InceptionV4~\cite{szegedy2017inception}, which performs avg/max pooling and different convolution operations in parallel before aggregating them as final representation.
\end{itemize}

\subsection{Result on SST}
We evaluate the optimal result architecture by training it from scratch and searching for the best hyper-parameters. We set batch size as 128, max input length as 64, hidden unit dimension for each layer as 256. Other hyper-parameters are optimized by grid search on the validation data (showed in the appendix). We compare our architecture with state-of-the-art networks designed by human experts, including 24-layers Transformer which is the text representation architecture leveraged in BERT~\cite{devlin2018bert}. We also compare to the original search spaces defined in ENAS~\cite{pham2018efficient}:

\begin{itemize}
    \item \textbf{ENAS-MACRO} is a macro search space over the convolutional and pooling layers, which is originally designed for image classification tasks. There are 6 operations in the search space: convolutions with filter sizes $3 \times 3$ and $5 \times 5$, depthwise-separable convolutions with filter sizes $3 \times 3$ and $5 \times 5$ \cite{chollet2017xception}, max pooling and average pooling of kernel size $3 \times 3$. In our experiments, we search for a macro neural network consisting of 24 layers.

    %\item \textbf{ENAS-MACRO + MORE OPTIONS} is an augmented search space that adding GRU and multi-head self-attention as additional operations. By this setting, we want to show the ablation study of two factors. First, we can see how much improvement can be obtained by capturing long-term dependencies in the architecture. Second, we demonstrate the benefit of multi-path ensemble by comparing the augmented search to TextNAS search space.

    \item \textbf{ENAS-MICRO} is a micro search space over normal and reduction cells.  There are two kinds of cells, i.e., normal cells and reduction cells. In each cell, there are $B=10$ nodes, where node 1 and node 2 are treated as the inputs of current cell. For each of the remaining $B-2$ nodes, the RNN controller makes two decisions: 1) selecting two previous nodes as inputs to the current node and 2) selecting two operations to apply on the input nodes. There are 5 available operations: identity, separable convolution with kernel size $3 \times 3$ and $5 \times 5$, average pooling and max pooling with kernel size $3 \times 3$. In our experiments, we stack the cells for 6 times. The normal cells and reduction cells are stacked alternatively.

\end{itemize}

We also compare to other search algorithms which have similar time complexities as ENAS, including DARTS~\cite{liu2018darts}, SMASH~\cite{brock2017smash}, One-Shot~\cite{bender2018understanding} and Random Search with Weight Sharing~\cite{li2019random}. Unless specified, we utilize the default settings of their open-source codes without tuning the hyper-parameters or modifying the proposed search spaces except for replacing all 2-D convolutions with 1-D (detailed settings can be found in the appendix).

%After grid search on the validation data, we set the initial learning rate as 0.005 and 0.08 respectively for \textit{SST} and \textit{SST-Binary} respectively.

\begin{table}[t]
    \caption{Results on SST dataset. For each dataset, we conduct significance test against the best reproducible model, and * means that the improvement is significant at 0.05 significance level.}
    \begin{center}
    \begin{small}
    \begin{sc}
    \begin{tabular}{l|c|c}
    \toprule
    Model & SST & SST-B \\
    \midrule
    %\cite{lai2015recurrent} & 47.21 & - \\
    Lai ET AL., 2015 & 47.21 & - \\
    %\cite{zhou2015c} & 49.20 & 87.80\\
    Zhou ET AL., 2015 & 49.20 & 87.80\\
    %\cite{liu2016recurrent} & 49.60 & 87.90 \\
    Liu ET AL., 2016 & 49.60 & 87.90 \\
    %\cite{tai2015improved} & 51.00 & 88.00 \\
    Tai ET AL., 2016 & 51.00 & 88.00 \\
    %\cite{kumar2016ask} & \textbf{52.10} & \textbf{88.60} \\
    Kumar ET AL., 2016 & \textbf{52.10} & \textbf{88.60} \\
    % \cite{vaswani2017attention} & 49.37& 86.66 \\
    24-layers Transformer & 49.37& 86.66 \\
    \midrule
    % ENAS-macro & \textbf{51.87} & \textbf{89.03} \\
    ENAS-macro & 51.55 &  \textbf{88.90} \\
    ENAS-micro & 47.00& 87.52 \\
    % DARTS & \textbf{51.95} & 86.33 \\
    DARTS & \textbf{51.65} & 87.12 \\
    SMASH & 46.65 & 85.94 \\
    % One-Shot & 50.39 & 89.45 \\
    One-Shot & 50.37 & 87.08 \\
    Random Search & 49.20& 87.15 \\
    \midrule
    TextNAS & \textbf{52.51} & $\textbf{90.33}^*$ \\
    % ARC-I& 51.43 & \textbf{89.44} \\
    % ARC-II &  50.67 & 88.87\\
    % ARC-III & \textbf{51.64} & 88.88 \\
    % ARC-I & 51.77 & \textbf{89.94} \\
    % ARC-II &  52.51 & 88.92 \\
    % ARC-III & \textbf{52.79} & 89.27 \\
    % \midrule
    % Joint Architecture & \textbf{53.44} & \textbf{90.23} \\
    \bottomrule
    \end{tabular}
    \end{sc}
    \end{small}
    \end{center}
    \vskip -0.1in

\label{table:SST}
\end{table}

The evaluation results are shown in Table \ref{table:SST}. We can see that the neural architecture discovered by TextNAS achieves competitive performances compared with state-of-the-art manual architectures, including the 24-layers Transformer adopted by BERT. At the same time, it outperforms other network architectures discovered automatically by other search spaces and algorithms. Specifically, the accuracy is improved by 11.7\% from ENAS-MICRO and 1.9\% from ENAS-MACRO on the SST dataset respectively, which shows the superiority of our novel search space for text representation. It should be noticed that there are other publications that have reported higher accuracies. However, they are not directly comparable to our scenario since they incorporate various kinds of external knowledge, e.g., BERT \cite{devlin2018bert} pre-trains on a large external corpus and \cite{yu2017refining} exploits syntax information in the Tree-LSTM model.

\subsection{Result on Architecture Transfer}
\label{exp:transfer}

\subsubsection{Text Classification}
We transfer the derived architecture as text representation networks to other eight text classification datasets\footnote{The datasets are available at http://xzh.me/} \cite{zhang2015character}. These datasets are from various domains including sentiment analysis, Wikipedia article categorization, news categorization and topic classification. The counts of samples are widely spread from hundreds of thousands to several millions as summarized in Table \ref{table:dataset}.

We follow the train/test split of the original datasets in all our experiments. For those datasets without validation set, we randomly select 5\% samples from the training set as validation data. For all datasets, we use pre-trained GloVe embedding to initialize word vectors and fine-tune them during training. To simplify the learning rate fine-tuning procedure for different datasets, we adopt an auto-decay strategy instead of cosine annealing. Given an initial learning rate, we use a small learning rate ($0.1 \times init\_rate$) to warm up the training procedure for 5 epochs; then we start from $init\_rate$ and decay it with a factor of 0.2 when the average validation accuracy of 7 recent epochs on the validation data drops. Finally, after 4 times of decay, we update the model for another 6 epochs on the full training set (training + validation). As a result, only one hyper-parameter, i.e., $init\_rate$, is required for each dataset. For critical hyper-parameters, we employ grid search on the validation data. Specifically, we search in $\{0.08, 0.05, 0.02\}$ for learning rate, $\{64, 128\}$ for batch size,  $\{64, 256, 512\}$ for max input length, $\{2 \times 10^{-9}, 2 \times 10^{-7}, 1 \times 10^{-6}, 2 \times 10^{-6}\}$ for $L_2$ regularization, $\{0.0, 0.2, 0.5\}$ for drop-out ratio, and $\{32, 64, 128, 256\}$ for hidden units dimension respectively. We observe that the Adam optimizer is not stable in several settings, so we adopt stochastic gradient descent with momentum 0.9 for training on all the datasets. More detailed settings are described in the appendix.

\begin{table*}[t]
    \caption{Test accuracy on the text classification datasets. For each dataset, we conduct significance test against the best reproducible model, and * means that the improvement is significant at 0.05 significance level.}
    \begin{center}
    \begin{footnotesize}
    \begin{sc}
    \begin{tabular}{l|c|c|c|c|c|c|c|c}
    \toprule
    Model& AG & Sogou & DBP& Yelp-B & Yelp & Yahoo & Amz & Amz-B \\
    \midrule
    % \cite{zhang2015character}    &         92.36&    \textbf{97.19}&    98.69&    95.64&    62.05&     71.20&   59.57&    95.07 \\
    Zhang ET AL., 2015    &         92.36&    \textbf{97.19}&    98.69&    95.64&    62.05&     71.20&   59.57&    95.07 \\
    %\cite{joulin2016bag} &         92.50&    96.80&    98.60&    95.70&    63.90&     72.30&  60.20&    94.60  \\
    Joulin ET AL., 2016 &         \textbf{92.50} &    96.80&    98.60&    95.70&    63.90&     72.30&  60.20&    94.60  \\
    % \cite{conneau2016very} &         91.33&    96.82&    98.71&     95.72 &   64.72&    73.43&     63.00&   95.72 \\
    Conneau ET AL., 2016 &         91.33&    96.82&    98.71&     \textbf{95.72} &  \textbf{64.72} &    \textbf{73.43} &     \textbf{63.00} &   \textbf{95.72} \\
    24-Layers Transformer &     92.17&   94.65&   \textbf{98.77} &   94.07&    61.22&    72.67&     62.65&  95.59  \\
    %\midrule
    %ENAS-macro & & & & & & & & \\
    %ENAS-micro & & & & & & & & \\
    \midrule
       ENAS-macro &   92.39&  96.79&  \textbf{99.01}&  \textbf{96.07}& 64.60&  73.16&  62.64& \textbf{95.80} \\
       ENAS-micro &   92.27&   \textbf{97.24}&  99.00 &  96.01&  64.72& 70.63&   58.27&  94.89  \\
       DARTS &  92.24&   97.18&   98.90&   95.84& 65.12& 73.12&   62.06&   95.48  \\
       SMASH &  90.88&  96.72&  98.86&  95.62&  \textbf{65.26}&  \textbf{73.63}&  \textbf{62.72}& 95.58  \\
       One-Shot &  92.06&  96.92&  98.89&  95.78& 64.78& 73.20&   61.30&   95.20  \\
       Random Search & \textbf{92.54}& 97.13& 98.98&  96.00& 65.23& 72.47&  60.91&  94.87  \\
       \midrule
    textnas  &   \textbf{93.14}& \textbf{96.76} & \textbf{99.01}&  $\textbf{96.41}^*$&    $\textbf{66.56}^*$&  $\textbf{73.97}^*$&    $\textbf{63.14}^*$&    $\textbf{95.94}^*$ \\
    % ARC-I  &   \textbf{93.14}& 96.76& \textbf{99.01}&  \textbf{96.41}&    \textbf{66.56}&    73.97&    63.14&    \textbf{95.94} \\
    % ARC-II     &   92.80&    \textbf{97.17}&    98.96&    96.22&    66.02&    \textbf{74.09}&    62.95&    95.90  \\
    % ARC-III &   91.71&    96.60&    98.96&    96.13&    66.26&    73.79&    \textbf{63.17}&    95.93 \\
    % \midrule
    % Joint Architecture &   \textbf{93.29}&    \textbf{97.35}&   \textbf{99.06}&    \textbf{96.61}&     \textbf{67.28}&    \textbf{74.57}&     \textbf{63.86}&   \textbf{96.25} \\
    \bottomrule
    \end{tabular}
    \end{sc}
    \end{footnotesize}
    \end{center}
    \vskip -0.1in
\label{table:transfer}
\end{table*}

The test accuracies on all datasets are shown in Table \ref{table:transfer}. The results demonstrate that the TextNAS model outperforms state-of-the-art methods on all text classification datasets except Sogou. One potential reason is that Sogou is a dataset in Chinese language, while the Glove embedding vectors are trained by English corpus. One can improve the performance by adding Chinese-language embeddings or char-embeddings, but we do not add them to keep the solution neat. In addition, we can pay a specific attention to the comparison of TextNAS with 29-layers CNN (Conneau ET AL., 2016) and 24-layers Transformer (VASWANI ET AL., 2017). As shown in the table, the TextNAS network improves two baselines by a large margin, indicating the advantage for mixture of different layers.

\subsubsection{Natural Language Inference}
%To further verify the versatility of our methodology,
We carry out experiments on two Natural Language Inference (NLI) datasets by leveraging the network architecture of TextNAS as sentence encoder. The \textit{SNLI} dataset\footnote{https://nlp.stanford.edu/projects/snli/}~\cite{bowman2015large} consists of 549,367 samples for training, 9,842 samples for validation and 9,824 samples for testing. The \textit{MultiNLI} dataset\footnote{https://www.nyu.edu/projects/bowman/multinli/}~\cite{N18-1101} contains 392,702 pairs for training. It has two separate sets for evaluation: \textit{MNLI-M} (matched set) has 9,815 pairs for validation and 9,796 pairs for testing; \textit{MNLI-MM} (mismatched set) contains 9,832 pairs for validation and 9,847 pairs for testing. Each sample is labeled with one of three labels: entailment, contradiction and neutral.

%\textit{SNLI} (Stanford Natural Language Inference)\footnote{https://nlp.stanford.edu/projects/snli/} is a commonly used open dataset for natural language inference task, which consists of 549,367 samples for training, 9,842 samples for validation and 9,824 samples for testing. \textit{MultiNLI} (Multi-Genre Natural Language Inference)\footnote{} is another prevailing NLI dataset. It contains 433k human-written sentence pairs from ten distinct genres, but only half of these genres are used in the training set (392,702 pairs). The matched validation (9,815 pairs) and test (9,796 pairs) sets include the same five genres as training data, while the mismatched validation (9,832 pairs) and test (9,847 pairs) set include all genres presented in the corpus. Each sample of these two datasets is labeled with one of the three labels: entailment, contradiction and neutral.

We initialize the word embedding layer by the concatenation of pre-trained GloVe embeddings and charNgram embeddings~\cite{hashimoto2016joint}. The word embedding vectors are fine-tuned during training. The outputs of all layers in the sentence encoder are linearly combined to produce the vector-based representation. We set the dimension of hidden units as 512 for all layers in the sentence encoder and 2400 for the fully connected layers before softmax output. Dropout is adopted on the output of each word-embedding, GRU and fully connected layer. Adam optimizer with learning rate decay strategy of cosine annealing is utilized to train the model. Detailed settings are optimized by grid search and presented in the appendix.

The evaluation results are illustrated in Table \ref{table:SNLI}. To get a fair comparison, we only compare with state-of-the-art sentence vector-based models that perform classification on the sole basis of a pair of fixed-size sentence representations. As shown in the table, TextNAS achieves competitive test accuracy on both \textit{SNLI} and \textit{MNLI} datasets consistently. In addition, it performs much better than the 24-layer Transformer, which verifies the effectiveness of our search space and methodology.

\begin{table}[t]
    \caption{Results on NLI datasets. For each dataset, we conduct significance test against the best reproducible model, and * means that the improvement is significant at 0.05 significance level.}
    \begin{center}
    \begin{small}
    \begin{sc}
    \begin{tabular}{l|c|c}
    \toprule
    \multicolumn{1}{c|}{Model} & SNLI & MNLI-m/mm \\
    \midrule
    % \cite{nie2017shortcut}  & 86.0 & 74.6 / 73.6 \\
    Nie and Bansal, 2017  & 86.0 & \textbf{74.6} / 73.6 \\
    % \cite{im2017distance}  & 86.3 &  74.1 / 72.9 \\
    Im and Cho, 2017 & 86.3 &  74.1 / 72.9 \\
    % \cite{talman2018hbmp} & 86.6 &  73.7 / 73.0 \\
    Talman ET AL., 2018 & 86.6 &  73.7 / 73.0 \\
    %\cite{chen2018enhancing}  & 86.6 & \textbf{73.8} / \textbf{74.0} \\
    Chen ET AL., 2018 & 86.6 & 73.8 / \textbf{74.0} \\
    %\cite{kiela2018dynamic}  & 86.7 & \textbf{75.0$^*$} / \textbf{74.2$^*$} \\
    %\cite{kiela2018dynamic}  & 86.7 & - \\
    Kiela ET AL., 2018 & \textbf{86.7} & - \\
    %\cite{yoon2018dynamic}  & \textbf{87.4} & - \\
    %Yoon ET AL., 2018  & \textbf{87.4} & - \\
    24-Layers Transformer & 85.2 & 70.4 / 70.2 \\
    %\midrule
    %ENAS-macro & & \\
    %ENAS-micro & & \\
    \midrule
    TextNAS & $\textbf{87.4}^*$ & \textbf{74.9} / \textbf{74.2} \\

    % ARC-I & \textbf{87.4} & \textbf{74.9} / \textbf{74.2} \\
    % ARC-II & 86.6 & 73.7 / 73.4\\
    % ARC-III & 87.1 & 74.1 / 73.0\\
    % \midrule
    % Joint Architecture & \textbf{87.8} & \textbf{75.8} / \textbf{75.2}\\
    \bottomrule
    \end{tabular}
    \end{sc}
    \end{small}
    \end{center}
    \vskip -0.1in
\label{table:SNLI}
\end{table}

To conclude, TextNAS generates novel and transferable network architecture for text classification and natural language inference tasks. By searching neural architectures on a relatively small dataset and then transferring it to larger ones, the network design procedure can be performed efficiently and effectively.

\section{Conclusion \& Future Work}
In this paper, we propose a novel architecture search space specialized for text representation by leveraging multi-path ensemble and a mixture of convolutional, recurrent, pooling, and self-attention layers. We demonstrate that by applying an efficient search algorithm, the TextNAS neural network architecture achieves state-of-the-art performance in various text-related applications. In addition, the architecture is explainable and transferable to other tasks. Future work mainly falls into three aspects: (1) uniting neural architecture search with state-of-the-art transfer learning frameworks, e.g., BERT; (2) exploring search acceleration techniques and conduct neural architecture search on larger datasets; (3) applying the TextNAS framework to other text-related tasks, such as Q\&A, machine translation and search relevance.

\footnotesize{\bibliography{aaai20}}
\bibliographystyle{aaai}

\appendix

%\section{Automatically Discovered Architectures}

% We visualize the architecture of ARC-II and ARC-III in Figure \ref{fig:ARC-II_ARC-III}. ARC-II and ARC-III are both 24-layer multi-path neural networks models, which are depicted below:

%\begin{itemize}
%    \item \textbf{ARC-II} consists of 11 convolution layers, 4 max-pooling layers, 5 average-pooling layers, 1 bidirectional GRU layer and 3 self-attention layers.
%    \item \textbf{ARC-III} consists of 8 convolution layers, 4 max-pooling layers, 8 average-pooling layers, 3 bidirectional GRU layers and 1 self-attention layer.
%\end{itemize}

\section{Neural Architecture Search Baselines}
State-of-the-art neural architecture search methods are mostly designed for image classification. In our experiments, we replace all 2-D convolutional operations with 1-D when applied to text-related applications. The detailed introduction and configuration of baseline methods are described as follows. Unless specified, we use the default hyper-parameters in the their open source codes\footnote{ENAS: https://github.com/melodyguan/enas \\ DARTS: https://github.com/quark0/darts \\ SMASH: https://github.com/ajbrock/SMASH \\ One-Shot: revised from DARTS \\ Random Search:  https://github.com/liamcli/randomNAS\_release}. For all experiments, we adopt learning rate decay with cosine annealing, where we set $T$ as 10 and tune $\lambda_{max}$ and $\lambda_{min}$ separately for each experiment.

\begin{figure*}
       \centering
       \includegraphics[height=1.2\textwidth]{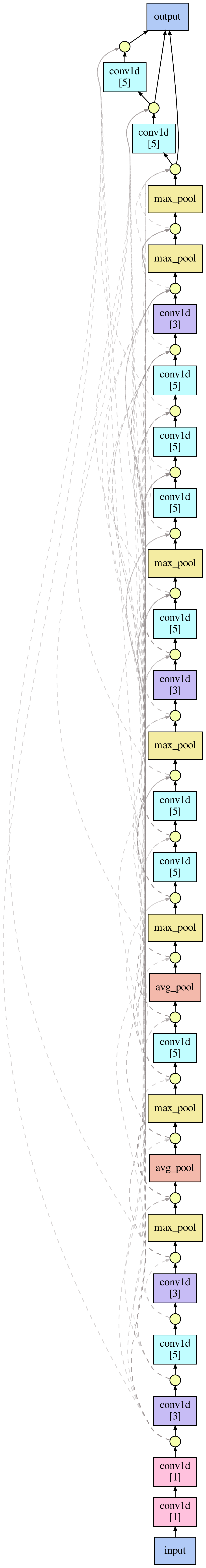}
       \caption{Visualization of architectures derived from \textit{ENAS-MACRO} search space: rectangles represent layers, circles represent summations, one-way arrows represent inputs, and dotted one-way arrows represent skip connections.}
\label{fig:macro-arc}
\end{figure*}

\textbf{ENAS-macro}~\cite{pham2018efficient} is a macro search space over the entire convolutional model, which is designed for image classification tasks. There are 6 operations in the search space: convolutions with filter sizes $3 \times 3$ and $5 \times 5$, depthwise-separable convolutions with filter sizes $3 \times 3$ and $5 \times 5$ \cite{chollet2017xception}, max pooling and average pooling of kernel size $3 \times 3$. In our experiments, we search for a macro neural network consisting of 24 layers. The architecture search results are visualized in Figure \ref{fig:macro-arc}.

\begin{figure*}
    \centering
    \subfloat[ENAS-MICRO-CELL]{
    \includegraphics[width=0.2 \textwidth]{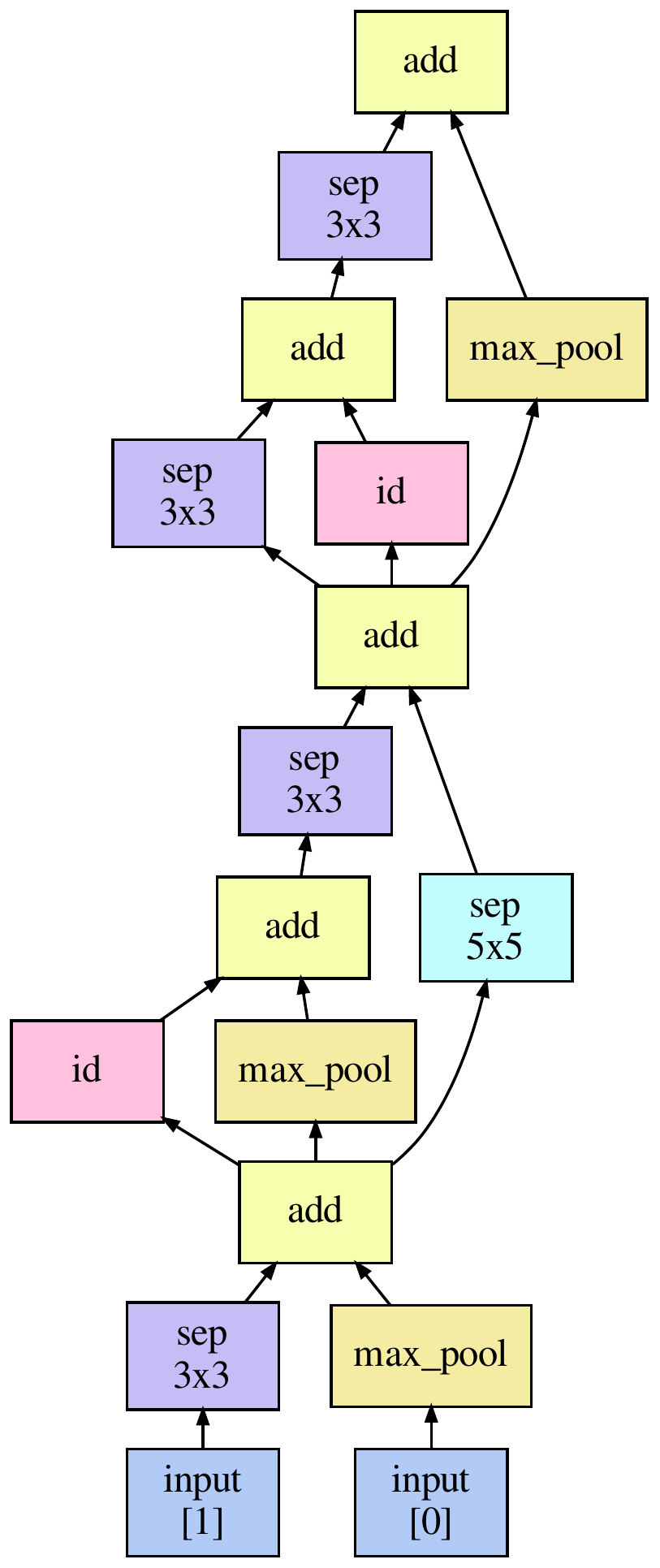}
    \label{fig:DAG_search_space}
    }
    \subfloat[ENAS-MICRO-REDUCTION]{
    \includegraphics[width=0.2 \textwidth]{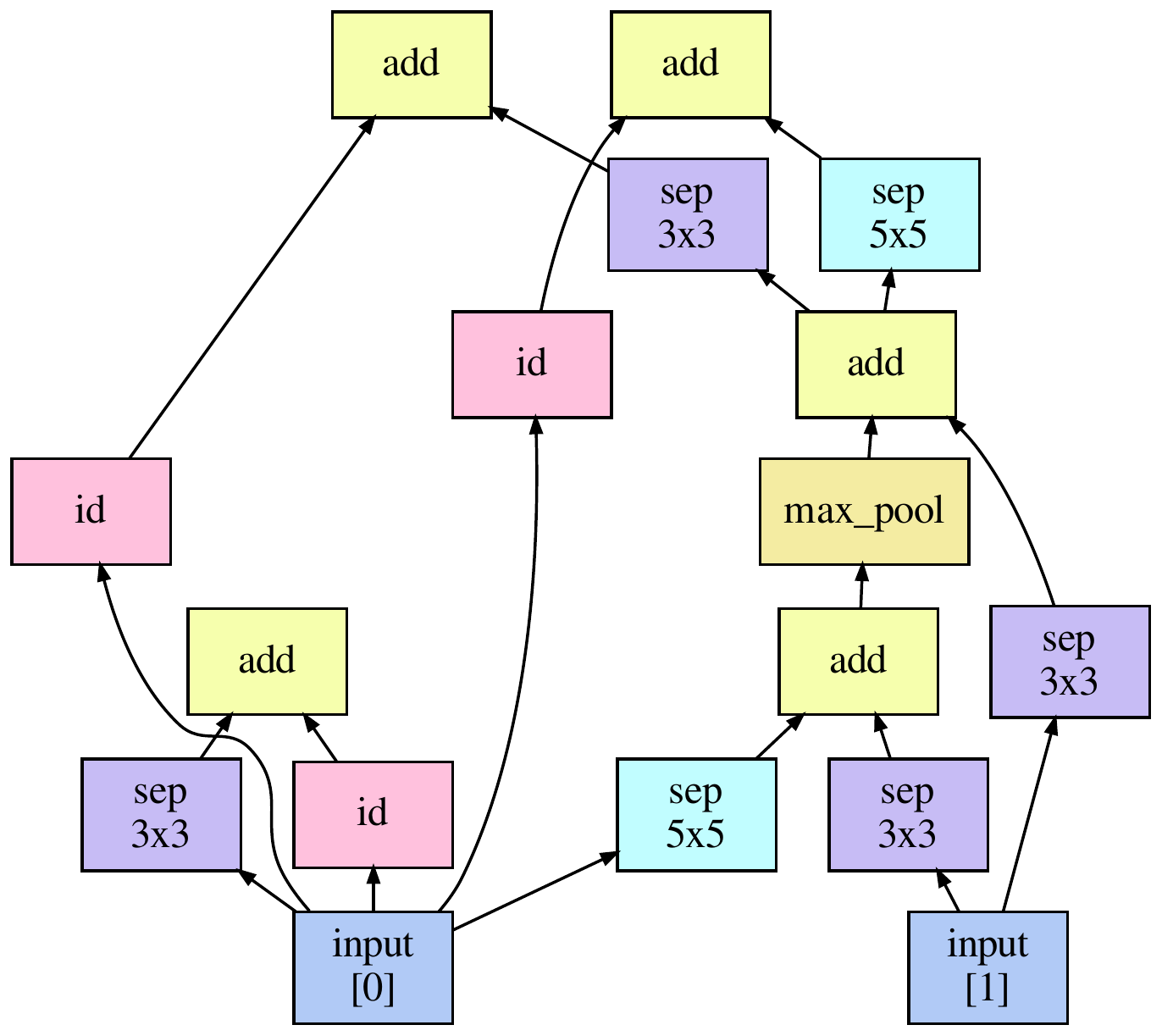}
     \label{fig:DAG_instance}
    }
    \caption{Visualization of architectures derived via \textit{ENAS-MICRO} search space: the architecture consists of two types of cells, the left figure for normal cell and the right figure for reduction cell. In all figures, the blue boxes represent for nodes and edges represent for operators.}
\label{fig:micro-arc}
\end{figure*}

\textbf{ENAS-micro}~\cite{pham2018efficient} is a micro search space over convolutional cells. There are two kinds of cells, i.e., normal cells and reduction cells. In a normal cell, there are $B=10$ nodes; node 1 and node 2 are treated as the cell’s inputs, which are the outputs of the two previous cells. For each of the remaining $B-2$ nodes, we ask the controller RNN to make two sets of decisions: 1) two previous nodes to be used as inputs to the current node and 2) two operations to apply to the two sampled nodes. The 5 available operations are: identity, separable convolution with kernel size $3 \times 3$ and $5 \times 5$, average pooling and max pooling with kernel size $3 \times 3$. The reduction cell can be constructed similarly by applying $stride=2$ for each operation, thus it reduces the spatial dimensions of its input by a factor of 2. In our experiments, we stack the cells for 6 times; and the normal cells and reduction cells are stacked alternatively. Concretely, the stack pattern is $normal -> reduction -> normal -> reduction -> normal -> reduction$. The result cells are visualized in Figure \ref{fig:micro-arc}.

\begin{figure*}
    \centering
    \subfloat[DARTS-CELL]{
            \includegraphics[height=0.18\textwidth, keepaspectratio=true]{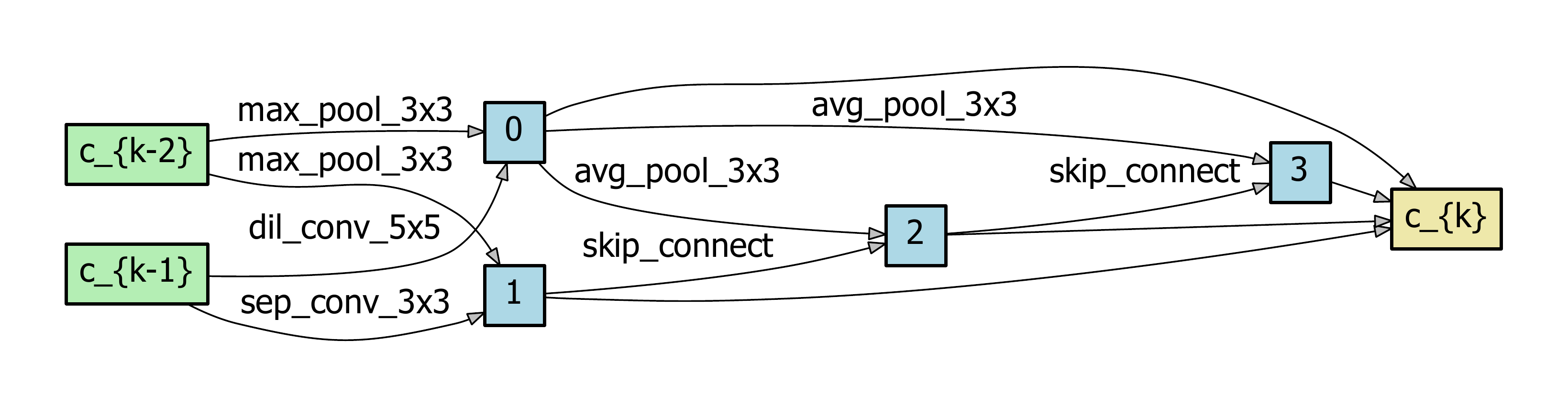}
    } \\
    \subfloat[DART-REDUCTION]{
            \includegraphics[height=0.18\textwidth, keepaspectratio=true]{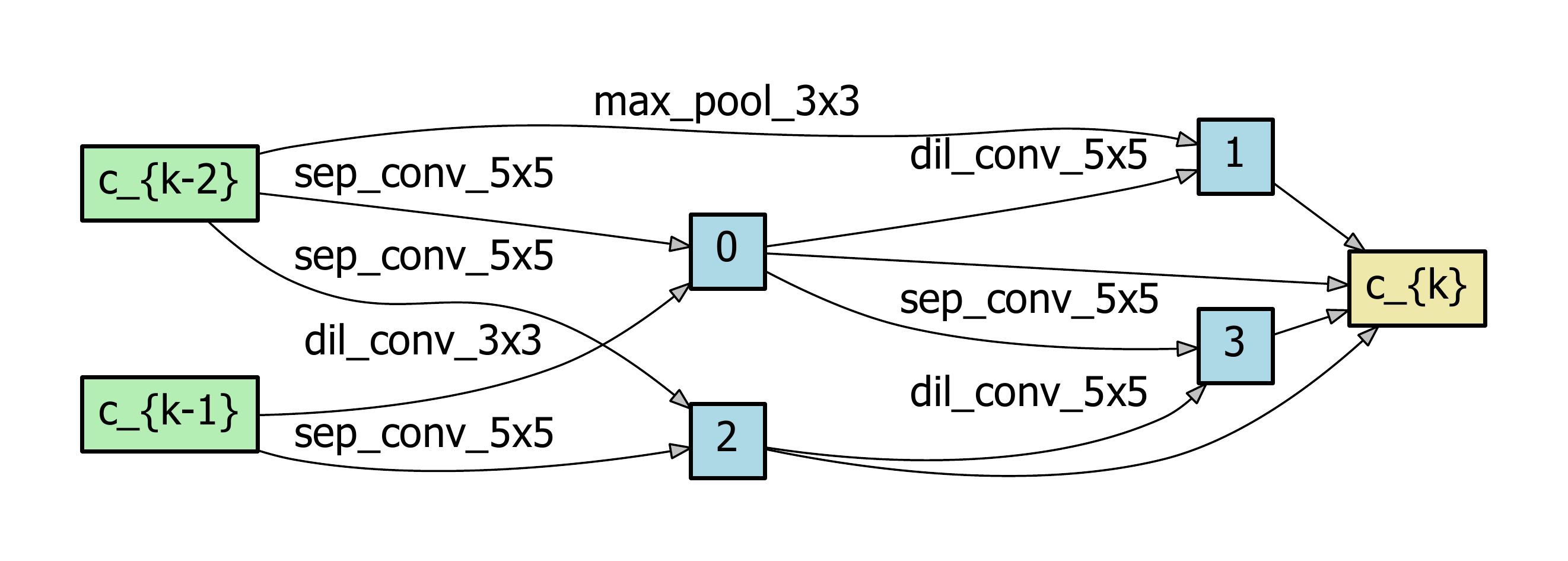}
    }
    \caption{Visualization of architectures derived via \textit{DARTS} search algorithm: the architecture consists of two types of cells, the upper figure for normal cell and the lower figure for reduction cell. In all figures, the blue boxes represent for nodes and edges represent for operators.}
\label{fig:darts-arc}
\end{figure*}

\textbf{DARTS}~\cite{liu2018darts} provides a continuous relaxation of the architecture representation, allowing efficient search of the architecture using gradient descent. The cell-based search space is the same as \textit{ENAS-micro}. Each layer is computed based on all of its predecessors, and the categorical choice of a particular operation is modeled by softmax over all possible operations. The task of architecture search is reduced to learning a set of continuous variables $\alpha = \{\alpha_{i, j, o}\}$, where $i$ and $j$ denote two arbitrary layers that satisfy $i<j$, and $o$ denotes a candidate operation. In our experiments, we stack the cells for 8 times in the search procedure, while in the evaluation procedure, we set the stack number for each dataset by grid search from 1 to 8. The result cells are visualized in Figure \ref{fig:darts-arc}.

%The space complexity is $O(KN!)$ where $K$ is the number of candidate operations and $N$ is the number of layers.

\begin{figure*}
    \centering
    \subfloat[SMASH]{
        \includegraphics[width=0.6\textwidth, height=1.2\textwidth]{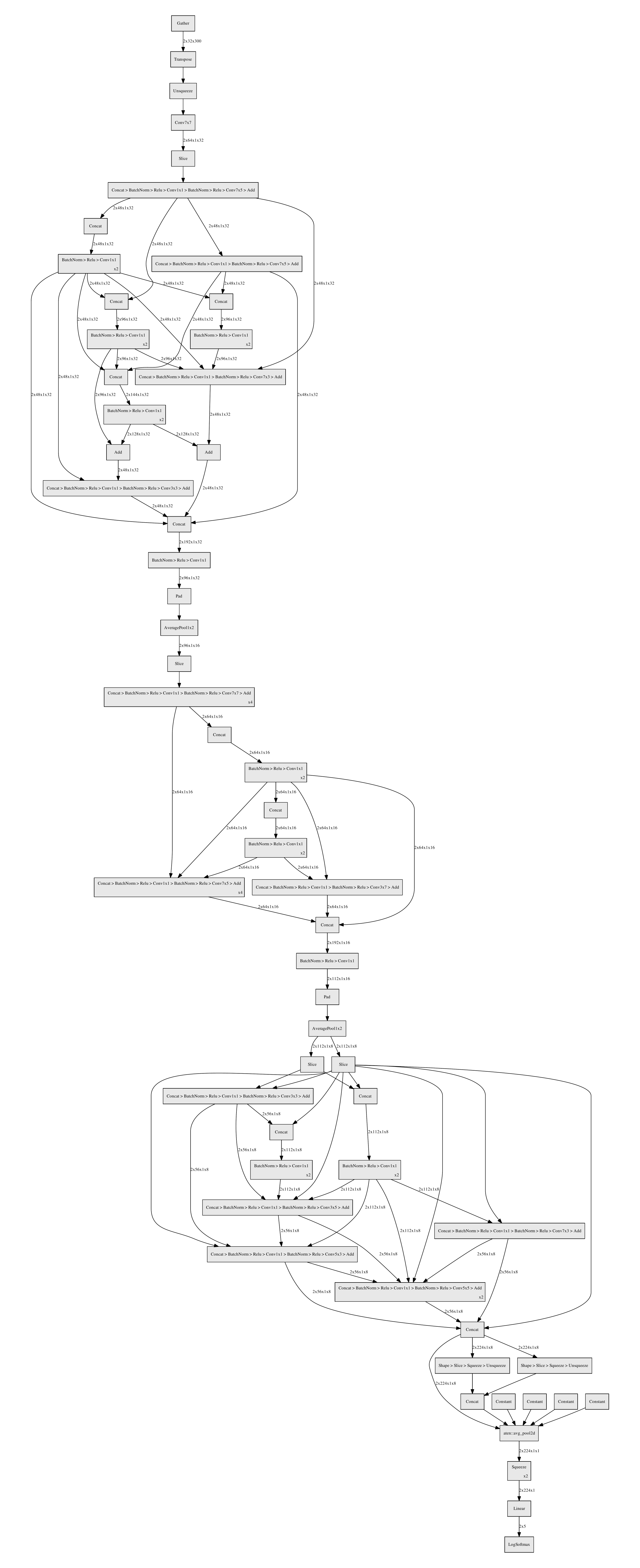}
    }
    \caption{Visualization of architectures derived from the \textit{SMASH} search space and algorithm: the block box represents for an operator or a group of operators.}
\label{fig:smash-arc}
\end{figure*}

\textbf{SMASH}~\cite{brock2017smash} bypasses the expensive procedure of fully training candidate models by using a hyper-network to dynamically generate the model weights. The search space is built by multiple blocks, where each block has a set of memory banks. When sampling an architecture, the number of banks and the number of channels per bank are randomly sampled at each block. A hyper-network is used to retrieve the model weights, while the evaluation results are leveraged to optimize the hyper-network. In our experiments, we set the widening factor to 4, depth value to 12, base channel number and maximum channel number to 8 and 64 respectively. The result architectures are visualized in Figure \ref{fig:smash-arc}.

\begin{figure*}
    \centering
    \subfloat[ONESHOT]{
      \includegraphics[height=0.35\textwidth]{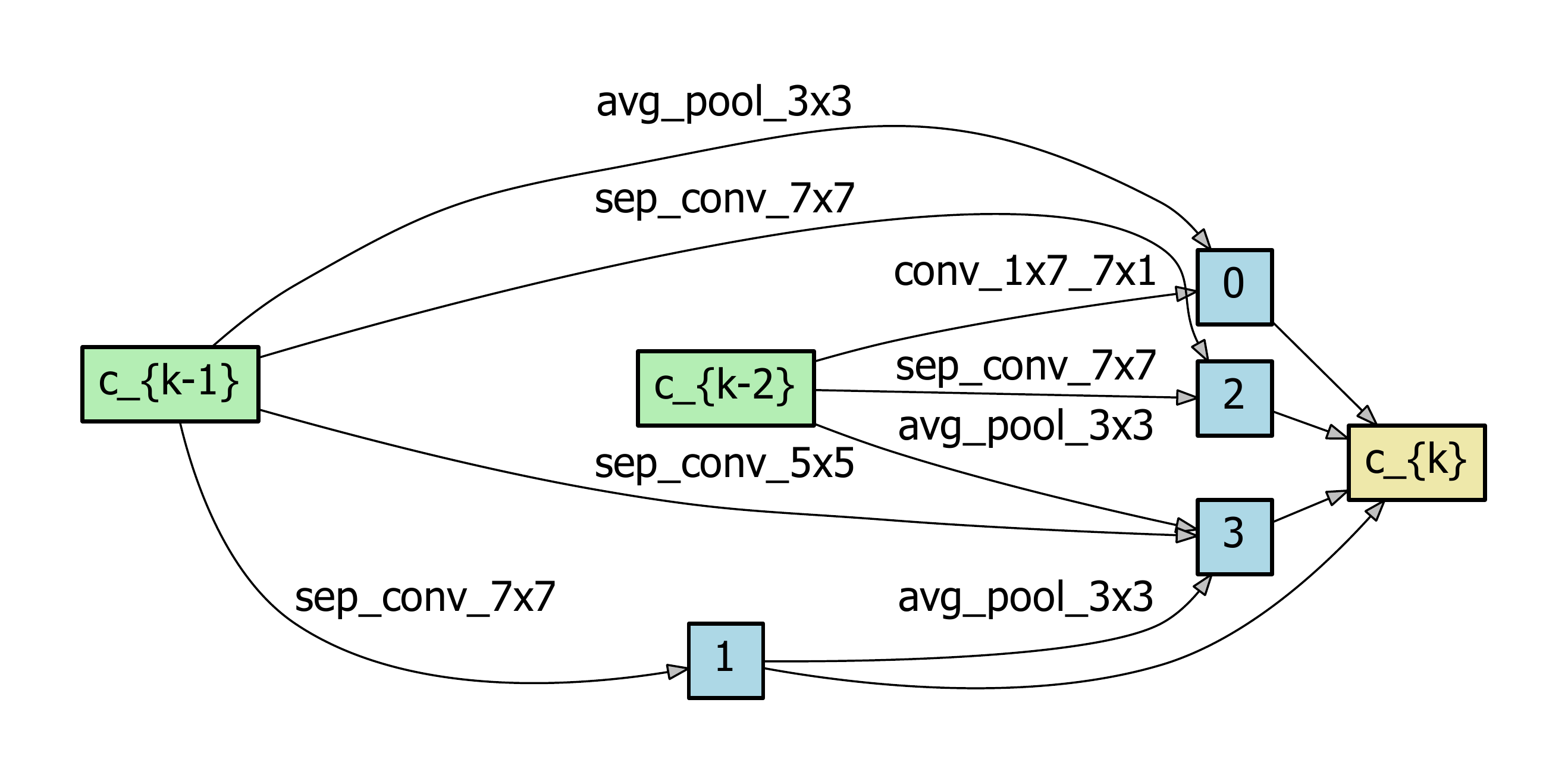}
    }
    \caption{Visualization of architectures derived via \textit{One-Shot} search algorithm: the blue boxes represent nodes and edges represent operators.}
\label{fig:oneshot-arc}
\end{figure*}

\textbf{One-Shot~}~\cite{bender2018understanding} shows that neither reinforcement learning controller nor hyper-networks are necessary for neural architecture search. It simply trains the one-shot model to make it predictive of the validation accuracies of the architectures. It then choose the architectures with the best validation accuracies and re-train them from scratch to evaluate their performance. Following the default setting, we stack the cells for 8 times in the search procedure, while in the evaluation procedure, we set the stack number for each dataset by grid search from 1 to 8. The result cells are visualized in Figure \ref{fig:oneshot-arc}.

\begin{figure*}
    \centering
    \subfloat[RANDOM\_SEARCH-CELL]{
            \includegraphics[height=0.2\textwidth, keepaspectratio=true]{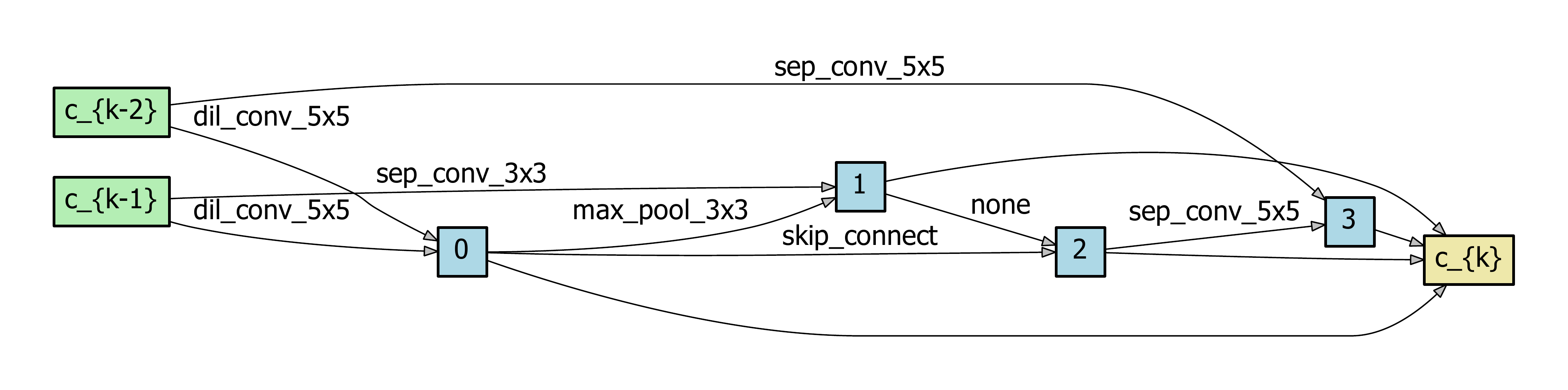} 
    }\\
    \subfloat[RANDOM\_SEARCh-REDUCTION]{
            \includegraphics[height=0.2\textwidth, keepaspectratio=true]{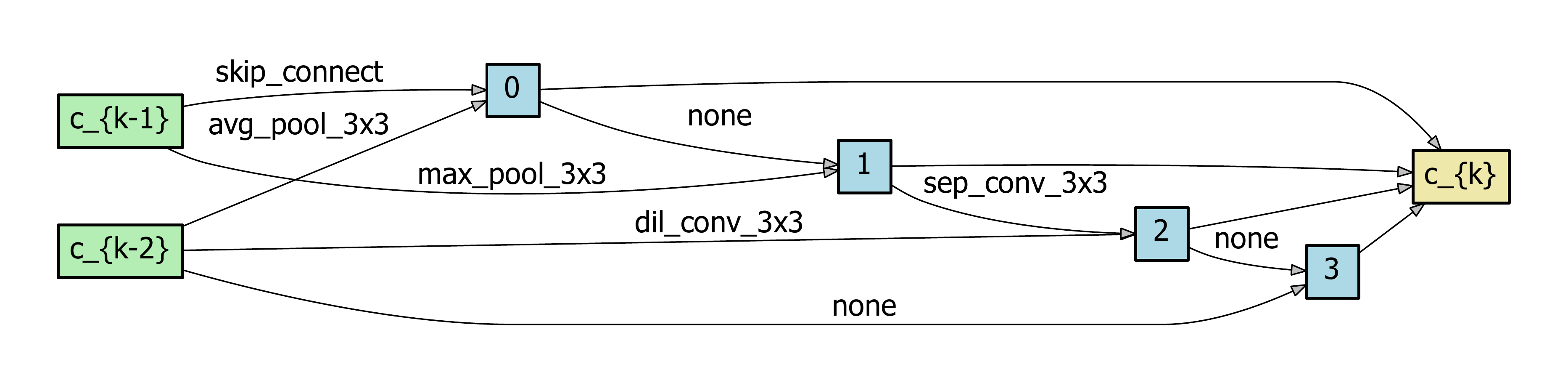}
    }
    \caption{Visualization of architectures derived via \textit{Random Search} algorithm: each architecture consists of two types of cells, the upper figure for normal cell and the lower figure for reduction cell. In all figures, the blue boxes represent for nodes and edges represent for operators.}
\label{fig:rws-arc}
\end{figure*}

\textbf{Random Search}~\cite{bergstra2012random} is a strong baseline for hyper-parameter tuning. In our experiments, we compare with the algorithm proposed by \cite{li2019random}. They treat NAS as a special hyper-parameter optimization problem and conduct random search with weight-sharing. The search space is the same as \textit{ENAS-micro} and \textit{DARTS}. We stack the cells for 8 times in the search procedure and employ grid search from 1 to 8 to find the best stack number for evaluation. The result cells are visualized in Figure \ref{fig:rws-arc}.

%The architectures discovered by these Neural Architecture Search baselines are depicted in Figure XX, including ENAS-micro, ENAS-macro, DARTS, SMASH, One-Shot and Random Search with Weight Sharing.
%The three architectures discovered in the single-path DAG search space is visualized in Figure \ref{fig:single-path}.

%\begin{figure*}
%\centering
%    \subfigure[ARC-II]{
%        \begin{minipage}[b]{0.9\textwidth}
%            \includegraphics[width=1\textwidth]{ARC-II} \\
%        \end{minipage}
%    }
%    \subfigure[ARC-III]{
%        \begin{minipage}[b]{0.9\textwidth}
%        \includegraphics[width=1\textwidth]{ARC-III} \\
%        \end{minipage}
%    }
%\caption{Visualization of multi-path architectures: rectangles represent layers, circles represent %summations, one-way arrows represent inputs, and dotted one-way arrows represent skip connections.}
%\label{fig:ARC-II_ARC-III}
%\end{figure*}

%\begin{figure*}
%\centering
%    \subfigure[ARC-single-path-I]{
%        \begin{minipage}[b]{0.3\textwidth}
%            \includegraphics[height=3.8\textwidth]{ARC-4} \\
%        \end{minipage}
%    }
%    \subfigure[ARC-single-path-II]{
%        \begin{minipage}[b]{0.3\textwidth}
%        \includegraphics[height=3.8\textwidth]{ARC-5} \\
%        \end{minipage}
%    }
%     \subfigure[ARC-single-path-III]{
%        \begin{minipage}[b]{0.3\textwidth}
%        \includegraphics[height=3.8\textwidth]{ARC-6} \\
%        \end{minipage}
%    }
%\caption{Visualization of single-path architectures: rectangles represent layers, circles represent summations, one-way arrows represent inputs, and dotted one-way arrows represent skip connections.}
%\label{fig:single-path}
%\end{figure*}

\section{Text Classification}

In all the experiments, we apply dropout (ratio=0.5) to the embedding layers, final output layers and self-attention layers. In addition, in the bidirectional GRU layers, we apply dropout (ratio=0.5) on the input and output tensors. Besides, for several time-consuming experiments, we employ \textit{sliding window} trick to accelerate the training procedure. Given a sentence, we utilize a sliding window to segment the long input sentence into several sub-sentences, where $window\_size$ and $stride$ are pre-defined hyper-parameters. The sub-sentences are fed separately to the neural network to output fixed-length vector representation for each sub-sentence. Then, a max pooling operator is applied on top to calculate the vector representation for the entire sentence. In all experiments using \textit{sliding window}, we set $window\_size$ as 64 and $stride$ as 32. Detailed settings of all experiments are listed in Table \ref{table:detailed_settings}.

%step size $step\_size$ to slide a fixed-size window over $s$. Then we can get a segmentation set $s^{'}=\{s^{'}_1,s^{'}_2, ..., s^{'}_n\}$, the length of $s^{'}_i$ is $w$ where $w$ is the window size. By this way, we cut a long sentence to multiple short sentences, and we feed $s^{'}$ instead of $s$ to the neural network. For $s^{'}_i$ we can get a corresponding output $x^{'}_i$, and for $s^{'}$ we get a output set $x^{'}=\{x^{'}_1,x^{'}_2, ..., x^{'}_n\}$, where $x^{'}_i$ is a fixed-size vector. Then we sum $x^{'}_1,x^{'}_2, ..., x^{'}_n$ together to get a fixed-size vector $x$. We use $x$ as the representation of $s$.

\begin{table*}[t]
    \small
    \caption{Detailed settings for experiments of text classification.}
    \begin{center}
    \begin{tabular}{c|l|c|c|c|c|c|c}
    \toprule
    Exp& batch size& max length& $l_2$& lr& sliding window& hidden size \\
    \midrule
    %\multirow{3}{*}{SST} & ARC-I& 128&  64& $1\times10^{-6}$& 0.005& no& 64 \\
    %       &      ARC-II&         128&  64& $2\times10^{-6}$& 0.005& no& 32 \\
    %       &      ARC-III&        128&  64& $1\times10^{-6}$& 0.005& no& 64 \\
    %\midrule
    %\multirow{3}{*}{SST-B} & ARC-I& 64& 64& $1\times10^{-6}$& 0.08& no& 32 \\
    %       &      ARC-II&        64& 64& $2\times10^{-6}$& 0.08& no& 32 \\
    %       &      ARC-III&       64& 64& $1\times10^{-6}$& 0.08& no& 32 \\
    %\midrule
    AG & 128& 256& $1\times10^{-6}$& 0.02& no& 256 \\
    %       &      ARC-II&        128& 256& $2\times10^{-6}$& 0.02& no& 256 \\
    %       &      ARC-III&       128& 256& $1\times10^{-6}$& 0.02& no& 256 \\
    \midrule
    Sogou & 64& 1024& $1\times10^{-6}$& 0.02& yes& 32 \\
     %      &      ARC-II&           64& 1024& $2\times10^{-9}$& 0.05& no& 64  \\
     %      &      ARC-III&          64& 1024& $1\times10^{-6}$& 0.02& yes& 32 \\
    \midrule
    DBP &  128& 256& $1\times10^{-6}$& 0.02& no& 64 \\
    %       &      ARC-II&         128& 256& $2\times10^{-6}$& 0.02& no& 64  \\
    %       &      ARC-III&        128& 256& $1\times10^{-6}$& 0.02& no& 64 \\
    \midrule
    Yelp-B & 128& 512& $1\times10^{-6}$& 0.02& no& 64 \\
    %       &      ARC-II&            128& 512& $2\times10^{-6}$& 0.02& no& 64  \\
    %       &      ARC-III&           128& 512& $1\times10^{-6}$& 0.02& no& 64 \\
    \midrule
    Yelp &  128& 512& $1\times10^{-6}$& 0.02& no& 64 \\
    %       &      ARC-II&           128& 512& $2\times10^{-6}$& 0.02& no& 64  \\
    %       &      ARC-III&          128& 512& $1\times10^{-6}$& 0.02& no& 64 \\
    \midrule
    Yahoo & 64& 1024& $1\times10^{-6}$& 0.02& yes& 32 \\
    %       &      ARC-II&           128& 512& $2\times10^{-7}$& 0.02& no&  64  \\
    %       &      ARC-III&          64& 1024& $1\times10^{-6}$& 0.02& yes& 32 \\
    \midrule
    Amz &  128& 256& $1\times10^{-6}$& 0.02& yes& 128 \\
    %       &      ARC-II&          128& 256& $2\times10^{-6}$& 0.02& yes& 128  \\
    %       &      ARC-III&         128& 256& $1\times10^{-6}$& 0.02& yes& 128 \\
    \midrule
    Amz-B & 128& 256& $1\times10^{-6}$& 0.02& yes& 128 \\
    %       &      ARC-II&           128& 256& $2\times10^{-6}$& 0.02& yes& 128  \\
    %       &      ARC-III&          128& 256& $1\times10^{-6}$& 0.02& yes& 128 \\
    \bottomrule
    \end{tabular}
    \end{center}
    \vskip -0.1in
\label{table:detailed_settings}
\end{table*}

\begin{table*}[t]
    \small
    \caption{Detailed settings for experiments of natural language inference.}
    \vskip 0.15in
    \begin{center}
    \begin{tabular}{c|l|c|c|c|c|c}
    \toprule
    Exp & lr & training epoch & $l_2$& dropout rate& penalization\\
    \midrule
    SNLI & $2\times10^{-4}$& 8& $2\times10^{-2}$& 0.2& 0 \\
    %       &      ARC-II&        $2\times10^{-4}$& 8& $2\times10^{-2}$& 0.2& $1\times10^{-3}$ \\
    %       &      ARC-III&       $2\times10^{-4}$& 8& $2\times10^{-2}$& 0.2& $1\times10^{-3}$ \\
    \midrule
    MNLI & $1\times10^{-4}$& 20& $1\times10^{-2}$& 0.2& 0 \\
    %       &      ARC-II&        $2\times10^{-4}$& 30& $2\times10^{-2}$& 0.3& $1\times10^{-3}$ \\
    %       &      ARC-III&       $2\times10^{-4}$& 20& $1\times10^{-1}$& 0.3& $1\times10^{-3}$ \\
    \bottomrule
    \end{tabular}
    \end{center}
\label{table:detailed_nli_settings}
\end{table*}

\section{Natural Language Inference}

In NLI experiments, we evaluate the result model of TextNAS by training it from scratch. Different from text classification, we discover that the concatenation of GloVe \cite{pennington2014glove} and charNgram \cite{hashimoto2016joint} performs better than only GloVe to initialize word embedding vectors. We set the dimension of hidden units as 512 for all layers in the sentence encoder and 2400 for the three fully-connected layers before softmax output. All 24 layers in the sentence encoder are linearly combined to produce the ultimate sentence embedding vector. All convolutions follow an ordering of ReLU, convolution operation and batch normalization. We also employ layer normalization after the outputs of bidirectional GRU and self-attention layers. Dropout is adopted on the output of each word-embedding, GRU and fully-connected layer. We set batch size as 32 and max input length as 128. Adam optimizer with cosine decay of learning rate and warm up over the first epoch are utilized to train the model. Besides the standard cross-entropy loss, we add $l_2$ regularization and penalization term on parameter matrices of the multi-head attention pooling layer~\cite{chen2018enhancing}. The optimal hyper-parameter values are task-specific and architecture-specific, so we carry out grid search in the following range of possible values and the final configurations are reported in the Table \ref{table:detailed_nli_settings}.
\begin{itemize}
\item Learning rate: $1\times10^{-4}$, $2\times10^{-4}$, $3\times10^{-4}$
\item Training epochs: 8, 12, 16, 20, 30
\item $l_2$ regularization: $1\times10^{-2}$, $2\times10^{-2}$, $5\times10^{-2}$, $1\times10^{-1}$, $2\times10^{-1}$, $5\times10^{-1}$
\item Dropout ratio: 0.1, 0.2, 0.3
\item Penalization: 0, $1\times10^{-1}$, $1\times10^{-2}$, $1\times10^{-3}$
\end{itemize}

\end{document}